\documentclass{article}

\usepackage{microtype}
\usepackage{graphicx}
\usepackage{booktabs}
\usepackage{hyperref}

\usepackage{xcolor}
\usepackage{amsmath}
\usepackage{url}
\usepackage{float} 
\usepackage{enumitem}

\setlist{topsep=4pt, partopsep=0pt, itemsep=0pt, parsep=0pt,
         before={\setlength{\parskip}{0pt}}}

\def\preprintbuild{}

\ifdefined\preprintbuild
  \usepackage[accepted]{mlsys2026}
  \makeatletter
  \renewcommand{\Notice@String}{Preprint. Not peer reviewed and not published.}
  \makeatother
\else
  \usepackage{mlsys2026}
\fi

\newcommand{\affDaytona}{Daytona, San Francisco, California, USA}
\newcommand{\affWaterloo}{University of Waterloo, Ontario, Canada}
\newcommand{\authorEmail}{dgraviet@daytona.io}
\newcommand{\coauthorEmail}{tanmay.shah@uwaterloo.ca}

\mlsystitlerunning{Returning to development states}

\begin{document}

\twocolumn[
\mlsystitle{Returning to Development States: When Full Sandbox Snapshots Help Coding Agents}


\mlsyssetsymbol{equal}{*}

\begin{mlsysauthorlist}
\mlsysauthor{Daniel Thi Graviet}{daytona}
\mlsysauthor{Tanmay Shah}{uw}
\end{mlsysauthorlist}

\ifdefined\preprintbuild
\begin{center}
\small
\affDaytona\\
\affWaterloo\\
\texttt{\authorEmail}\quad\texttt{\coauthorEmail}
\end{center}
\fi

\mlsysaffiliation{daytona}{\affDaytona}
\mlsysaffiliation{uw}{\affWaterloo}

\mlsyscorrespondingauthor{Daniel Thi Graviet}{\authorEmail}

\mlsyskeywords{coding agents, test-time search, sandbox snapshots, Go-Explore, Terminal-Bench}

\vskip 0.3in

\begin{abstract}
Coding agents often make useful progress before they fail, but retry-from-scratch discards the machine state created along the way. This makes failure recovery an interesting search problem: if intermediate environment states preserve meaningful progress, restoring and branching from them could outperform repeatedly restarting from the initial state. We implement Go-Explore-style search over full Daytona snapshots and find that higher and more efficient solve rates depend mainly on progress that sits outside the repository, such as databases, services, and other runtime state. Our results suggest that full-snapshot search is most effective when long-horizon coding work spans multiple services whose state cannot be reconstructed from a single diff.
\end{abstract}
]

\printAffiliationsAndNotice{}  

\section{Introduction}
\label{sec:intro}

A coding-agent failure is rarely a clean reset point. Consider an agent debugging a repository. It installs missing dependencies, reproduces the failing test, modifies several files, and creates a local database needed by the application. It then makes one bad edit and fails the task. A standard retry starts again from the original repository, discarding not only the bad edit but also every useful change to the machine that came before it.

Most test-time scaling methods operate above this machine state. Independent retries sample new trajectories from the initial environment. Memory systems preserve text. Search methods preserve source-level state such as commits and file context. None of these necessarily preserve the complete execution environment, including installed packages, running services, generated artifacts, databases, caches, and other state accumulated during interaction.

Go-Explore~\citep{ecoffet2021goexplore} offers a natural alternative. It remembers promising states, returns to them, and explores again. We apply this idea to coding agents by treating a full Daytona sandbox snapshot as the searchable state. A snapshot freezes the live machine so that a new agent can resume from an intermediate point rather than reconstructing it from the repository or a textual summary.

We initially hypothesized that branching from promising snapshots would solve more tasks than independent retry under a fixed token budget. Our experiments refute that broad hypothesis on the Terminal-Bench~\citep{merrill2026terminalbench} tasks we tested, but they also reveal when environment-level state matters and what prevents snapshot search from being effective today.

This paper makes three contributions:
\begin{itemize}
    \item We provide a matched-budget comparison between snapshot branching and independent retry. On our pre-registered harder Terminal-Bench subset, retry solves 17/25 tasks while snapshot-based search solves 6/25, establishing that reliable restoration alone is not sufficient to improve test-time scaling (Section~\ref{sec:retry}).
    \item We identify concrete failure modes of environment-level search. Snapshot branching loses through budget fragmentation, selection of misleading intermediate states, agent handoff cost, and weak signals for deciding which states deserve further exploration (Section~\ref{sec:failure}).
    \item We show that the value of snapshots depends on what information the environment contains. On \texttt{extract-elf}, compressed source-level state can outperform a full snapshot, while on a custom task whose progress lives in a SQLite database outside git, full snapshots solve 4/5 attempts and source-only restoration solves 0/5 (Section~\ref{sec:state}). These results suggest that environment-level search is most useful when progress cannot be reconstructed from repository state alone.
\end{itemize}

\section{The Problem}
\label{sec:problem}

A coding-agent trajectory changes more than source files. During execution, the agent may install packages, generate artifacts, populate databases, modify configuration, start services, or otherwise alter the machine in ways that affect future actions. A retry from the original repository loses all of this state, while a source-level checkpoint such as a git diff preserves only the subset represented in files under version control. This creates a simple question. When an agent makes useful progress before failing, what state must we preserve to make that progress reusable? If source state is sufficient, simpler checkpoints suffice. But if critical progress lives in the execution environment itself, such as installed packages, service states, and runtime artifacts, full-machine snapshots become necessary. They uniquely preserve this state without the cost of re-execution, making them valuable for long-horizon tasks where such progress accumulates.

\section{The Idea}
\label{sec:idea}

The core idea is simple. Instead of treating a failed coding-agent run as one indivisible attempt, treat it as a path through a sequence of machine states. As the agent works, some states are more promising than others. Perhaps the bug has been reproduced, the right dependency is installed, or most of the fix is already in place. We periodically snapshot those states and store them in an archive. When a trajectory fails, we do not have to restart from the beginning. We can restore one of the archived states, place a new agent there, and continue exploring from that point. In this view, a sandbox snapshot plays the same role that a saved game does in search. It lets us return to a useful intermediate position and try a different continuation without replaying everything that came before. The important distinction is that the saved state is the entire execution environment, not just the agent's text history or a patch. This gives the search procedure access to any progress encoded in the live machine.

\begin{figure}[H]
    \centering
    \includegraphics[width=\columnwidth]{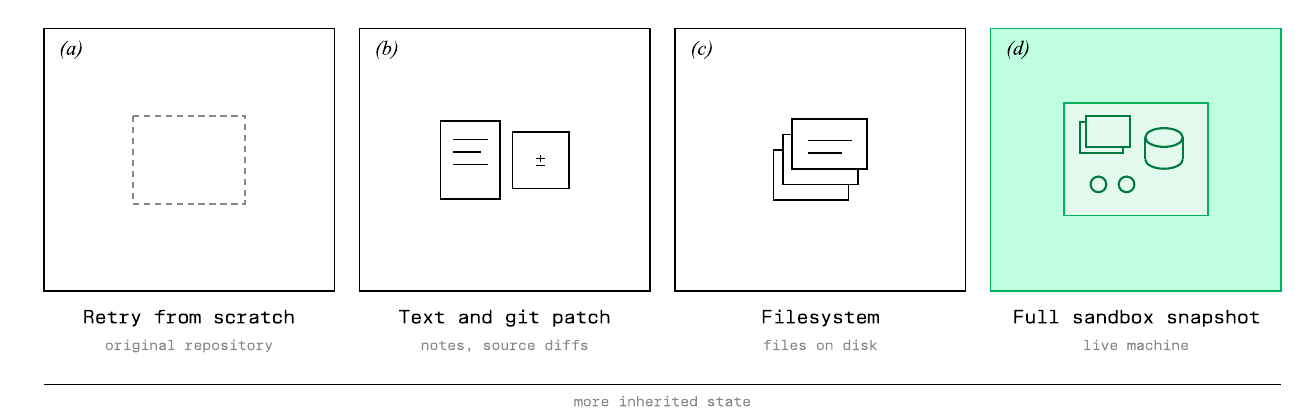}
    \caption{What the next attempt starts from. (a)~Retry from scratch uses the original repository. (b)~Text and git patches keep notes and source diffs. (c)~A filesystem checkpoint keeps files on disk. (d)~A full sandbox snapshot keeps the live machine. We study (d) as a restore primitive.}
    \label{fig:what-is-preserved}
\end{figure}

\section{System and Evaluation}
\label{sec:system}

We evaluate whether returning to intermediate machine states improves task success compared with spending the same inference budget on independent retries.

\subsection{Snapshot-based Search}
\label{sec:search}

Our system periodically saves promising points during a coding-agent run. For example, it may save a snapshot after the agent edits files, runs tests, or makes other useful progress.

Each snapshot captures the full Daytona sandbox at that moment. We store these snapshots in an archive and rank them based on simple signals such as file edits and test activity. Future selectors could adopt ideas from Intelligent Go-Explore, which uses a foundation model to pick which archived state to restore from~\citep{lu2025ige}.

In practice, when an agent fails on a hard task, restarting from the beginning loses all intermediate progress. Instead, our system identifies a promising saved state from the failed run, restores the sandbox to that point, and launches a new child trajectory. The child inherits the restored machine state but receives a fresh agent context, allowing a different approach from a promising midpoint.

This is the core Go-Explore idea. Instead of always restarting from the beginning, we return to a promising state and explore a different continuation from there.

We compare this against independent retry, where the same total token budget is spent on fresh attempts that all start from the original task state. Both methods use the same model and the same total inference budget. The main difference is whether later attempts start over or continue from saved progress.

\begin{figure}[H]
    \centering
    \includegraphics[width=\columnwidth]{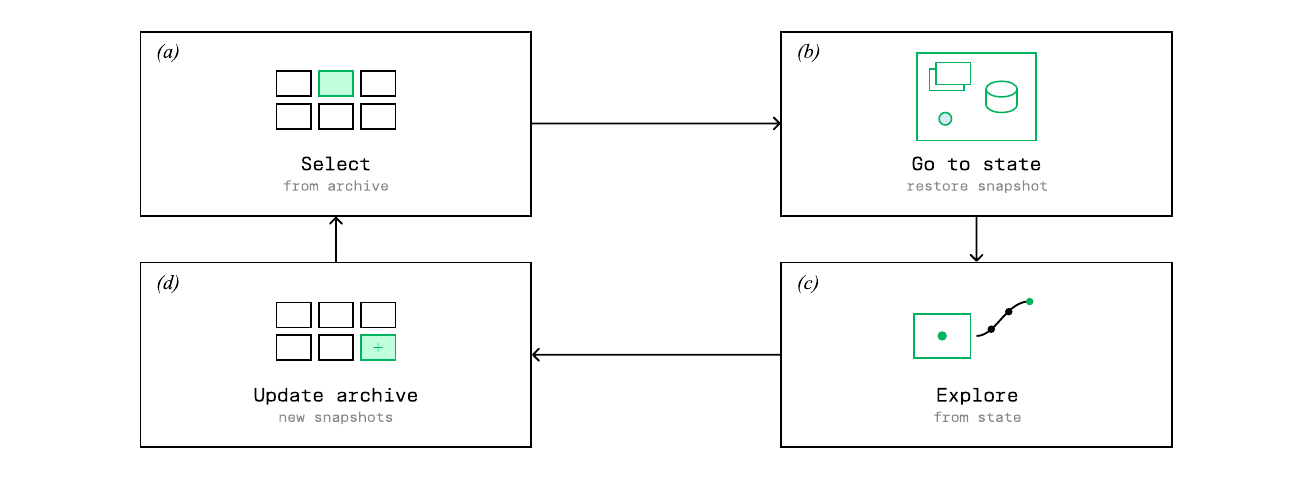}
    \caption{Search over Daytona sandboxes. (a)~Select a snapshot from the archive. (b)~Restore that sandbox (Go-Explore's ``go to state''). (c)~A new agent explores from the restored machine. (d)~New snapshots update the archive. Independent retry is the matched-budget baseline and is not shown.}
    \label{fig:search-pipeline}
\end{figure}

\subsection{Experimental Setup}
\label{sec:setup}

Unless otherwise noted, all experiments use Claude Haiku 4.5 on Terminal-Bench 2.0~\citep{merrill2026terminalbench} through Harbor~\citep{harbor2026} and Daytona.\footnote{\url{https://www.daytona.io/}} We intentionally use a lower-cost model because this work focuses on token-efficient agent search under fixed budgets. Models in this cost range are also practical for large numbers of rollouts in agent harnesses, where repeated attempts can quickly become expensive.

Our headline comparison uses five tasks selected for intermediate single-attempt difficulty. For each task, we run five seeds with a total budget of one million tokens per seed. Independent retry spends that budget across three fresh attempts, while snapshot search uses one root trajectory followed by up to two restored children. The task set and protocol were fixed before examining the final comparison.

\section{Snapshot Search vs Retry}
\label{sec:retry}

We test whether branching from promising intermediate states can outperform repeated fresh attempts under the same total token budget.

\begin{figure}[H]
    \centering
    \includegraphics[width=\columnwidth,height=1.2in,keepaspectratio]{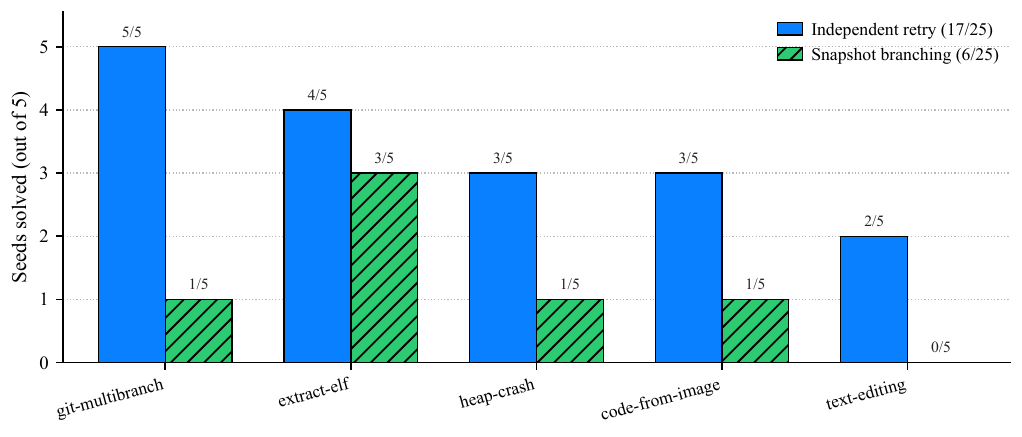}
    \caption{Retry 17/25 vs.\ snapshot branching 6/25 on the pre-registered set (five tasks $\times$ five seeds, 1M tokens).}
    \label{fig:headline}
\end{figure}

On this Terminal-Bench subset, independent retry outperforms snapshot branching on the selected tasks and solves more task-seed pairs overall.

\begin{table}[H]
    \centering
    \caption{Headline solve rates on the pre-registered task set (five seeds per task, 1{,}000{,}000 tokens per seed).}
    \label{tab:headline-results}
    \small
    \setlength{\tabcolsep}{4pt}
    \begin{tabular}{@{}lcc@{}}
        \toprule
        Task & Retry & Branch \\
        \midrule
        \texttt{git-multibranch}          & 5/5 & 1/5 \\
        \texttt{extract-elf}              & 4/5 & 3/5 \\
        \texttt{custom-memory-heap-crash} & 3/5 & 1/5 \\
        \texttt{code-from-image}          & 3/5 & 1/5 \\
        \texttt{large-scale-text-editing} & 2/5 & 0/5 \\
        \midrule
        Total                             & \textbf{17/25 (68\%)} & \textbf{6/25 (24\%)} \\
        \bottomrule
    \end{tabular}
\end{table}

However, snapshot search is not simply failing to restore useful computation. We observe four cases where the root trajectory fails but a child restored from that root later succeeds. Three occur on \texttt{extract-elf} and one on \texttt{custom-memory-heap-crash}. In a separate follow-up experiment, we specifically tested whether snapshot branching could recover from failed roots. On \texttt{extract-elf}, snapshot branching successfully recovered \textbf{67\%} of failed roots, substantially higher than independent retry's overall success rate of \textbf{38\%} on that task. On \texttt{regex-log}, the recovery rate was more modest at \textbf{13\%}. This shows that branching can salvage a failed root when a useful intermediate state exists. On the Terminal-Bench tasks we tested, those midpoints were not frequent enough for branching to win.

\section{What State Is Worth Preserving?}
\label{sec:state}

The headline experiment asks whether full snapshots improve search. A separate question is whether the full machine state itself contains information worth preserving.

We test this by comparing different ways of saving the same intermediate progress.

On tasks that can be easily reconstructed through git, saving only the changed files is often enough. On one of our tasks, \texttt{extract-elf}, children restored from a Git diff or from replayed commands succeeded on 4/4 seeds. This contrasts with children restored from full snapshots, which frequently ran out of budget for the reasons discussed in Section~\ref{sec:failure}. For this kind of task, most useful progress is already captured in the repository, so restoring the entire machine adds little.

We then construct \texttt{staged-service-repair}, where the agent's progress depends on a SQLite database at \texttt{/var/lib/inventory/store.db} that must exist for the verifier to pass. This reflects a common production scenario: agents modify system state such as databases, caches, and services that version control systems never track. A Git diff captures only code changes, not this underlying execution state, making full snapshots essential.

We freeze a Daytona snapshot partway through the task and compare three ways of starting the next agent, all with the same remaining token budget of 200{,}000:

\begin{itemize}
\item \textbf{Clean restart}: Start from the original task state with no prior progress.
\item \textbf{Git diff applied}: Restart with the saved Git diff restored, but without the database state.
\item \textbf{Full snapshot restore}: Restore the complete Daytona snapshot, including both code changes and database state.
\end{itemize}

\begin{figure}[H]
    \centering
    \includegraphics[width=\columnwidth]{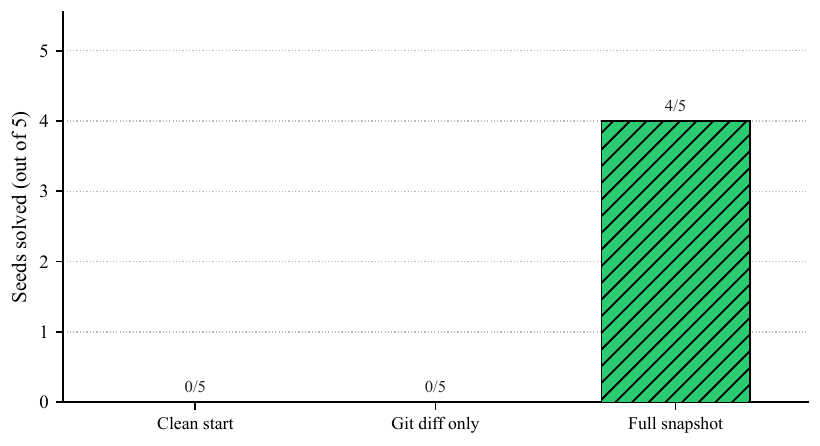}
    \caption{\texttt{staged-service-repair}, same planted checkpoint, 200{,}000 remaining tokens. The snapshot carries the warehouse database; the git patch does not. Clean 0/5, git diff 0/5, full snapshot 4/5. Not pooled with Figure~\ref{fig:headline}.}
    \label{fig:warehouse-arms}
\end{figure}

Here, preserving the environment changes the outcome. The Git diff does not contain the database state required by the verifier, while the snapshot does. Restoring the full machine state cuts the required inference nearly in half. Snapshot restorations solved the task with just 21k–41k tokens, while clean restarts and Git-diff starts consumed significantly more tokens and still failed.

\begin{figure}[H]
    \centering
    \includegraphics[width=\columnwidth]{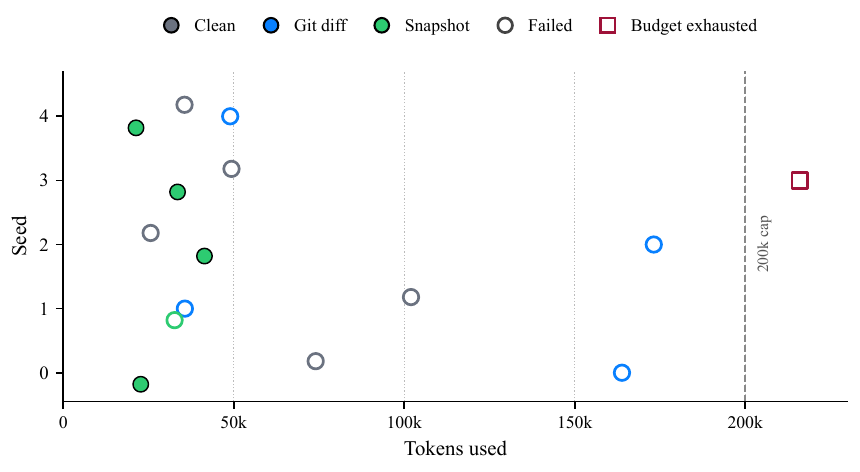}
    \caption{Token use at the same remaining cap. Filled markers solved. The square is a diff run that exhausted the 200k budget. Seed~1's snapshot (open green) restored the database and still failed.}
    \label{fig:warehouse-tokens}
\end{figure}

The one snapshot failure is not a restore failure. On seed~1, the warehouse database was present. The agent read the remaining migration, started the HTTP server, and marked the task complete without applying that migration. A new agent in a restored machine still has to notice what work remains.

Together, these experiments suggest a clear boundary for when full-machine restoration is most useful. It matters most when important progress lives outside the repository, such as in databases, installed packages, running services, generated artifacts, or other machine state that cannot be recovered from Git or source files alone.

\section{Failure Analysis}
\label{sec:failure}

The negative headline result is highly structured. We observe four recurring failure modes.

\begin{figure}[H]
    \centering
    \includegraphics[width=\columnwidth]{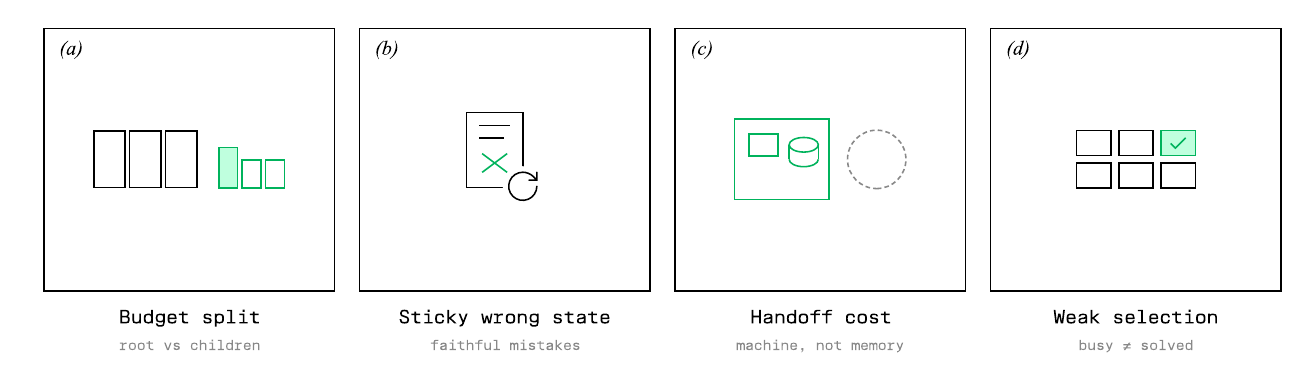}
    \caption{Why snapshot search still loses on file-centric tasks. (a)~Root and children share one token budget, so either can be underfunded. (b)~Restore is faithful to incorrect files. (c)~The child inherits the machine, not the parent's understanding. (d)~Easy signals of activity are not the verifier.}
    \label{fig:failure-modes}
\end{figure}

\paragraph{Budget fragmentation.}
A retry receives a complete independent attempt. Snapshot search instead spends part of its budget producing the root and divides the remainder among children. A weak root creates a weak archive, while short-lived children often run out of budget before completing a repair. Our current approach uses uniform budget allocation, but this is a natural target for improvement. Adaptive budgeting strategies, learned allocation policies, or task-specific heuristics could likely balance root exploration against child continuation more effectively. We expect future work to substantially improve upon this naive distribution.

\paragraph{Wrong-state attractors.}
Restoration faithfully preserves mistakes as well as progress. If the selected checkpoint contains a plausible but incorrect implementation, the child often continues editing around that decision rather than reconsidering it. Faithful restoration can therefore make an incorrect state unusually sticky. As described in Section~\ref{sec:search}, we rank snapshots with simple progress proxies rather than treating them uniformly. Those proxies are still weak, and better selection remains open. Related work such as Intelligent Go-Explore uses foundation models for this choice~\citep{lu2025ige}. State quality assessment, diversity mechanisms that encourage exploration of alternative solutions, or methods to detect and escape incorrect attractors could help children break out of bad local decisions.

\paragraph{Handoff cost.}
A fresh agent does not inherit the parent's understanding of the environment. It may receive the parent's filesystem but still need to rediscover how the repository works, how tests should be run, and what the previous agent was attempting. In one inspected trajectory, a child consumed roughly 190{,}000 tokens over 26 steps, while a successful fresh retry required about 90{,}000 tokens over 11 steps. This suggests an opportunity for better context transfer. State summaries, environmental documentation, or explicit handoff protocols could help new agents reuse the parent's implicit knowledge. Such mechanisms could substantially reduce the overhead of agent transitions.

\paragraph{Weak state selection.}
Deciding that a state is worth returning to is itself difficult. File edits, successful commands, or locally passing tests are imperfect proxies for actual progress. Early versions of our selector sometimes ranked package installation or claimed completion as strong evidence of progress, causing the archive to retain states that were easy to recognize but not useful to continue from. This points to a clear limitation in our ranking approach. Learned selection heuristics, task-specific progress metrics, or verifier feedback could improve the quality of archived states and help focus search on genuinely valuable checkpoints.

These failures point to a broader distinction. Snapshotting solves the state-restoration problem, but it does not solve the state-selection problem. Effective environment-level search requires both.

\section{Threats to Validity}
\label{sec:threats}

Our results cover one model tier and five screened Terminal-Bench tasks with five seeds each. The tasks were selected for intermediate baseline difficulty rather than sampled randomly, so the appropriate unit of generalization is the task rather than the individual trajectory.

Terminal-Bench is also largely file-centric. Tasks involving long-lived services, databases, caches, package installation, simulation state, or other persistent machine changes may place greater value on complete environment restoration.

Finally, we compare against independent retry rather than stronger search baselines such as best-of-N judging or SWE-Search, and we do not claim that the root/child budget allocation or snapshot-selection policy used here is optimal.

\section{Related Work}
\label{sec:related}

Our work sits at the intersection of test-time search for software agents, iterative program repair, and state-return methods from reinforcement learning. The main distinction we study is what gets preserved between attempts: text, source code, trajectories, or the complete execution environment.

\paragraph{Test-time scaling for software engineering.}
CodeMonkeys provides an inspiring demonstration of how much software agents can gain from additional test-time computation. It samples many editing trajectories, runs model-generated tests, and selects among candidate patches. The work shows that both serial and parallel test-time compute can substantially improve software-engineering performance~\citep{ehrlich2025codemonkeys}. SWE-Search takes a complementary approach by organizing search with Monte Carlo tree search, value estimation, and iterative feedback to decide which software states deserve more exploration~\citep{antoniades2024swesearch}.

Our work builds on this broader idea of spending computation more intelligently at test time. We focus on a different question. Instead of asking only how many trajectories to sample or how to score them, we ask what state a branch should resume from. In our system, a branch can inherit the complete live sandbox rather than reconstructing progress from repository state or agent context. Our results suggest that preserving more state is useful in some settings, but that good state selection and budget allocation are still essential.

\paragraph{Iterative repair and memory.}
RepairAgent is an important example of an autonomous repair system that can gather information, modify code, call tools, and validate candidate repairs through repeated interaction with a software environment~\citep{bouzenia2024repairagent}. Reflexion offers another influential approach to learning from failure. It converts feedback from previous attempts into textual reflections that guide future trials~\citep{shinn2023reflexion}.

These systems show that failed attempts can contain useful information that should not be discarded. Our work explores the same idea at the machine-state level. Textual memory or source patches may be enough when progress can be expressed in words or reconstructed from files. When progress lives in a database, installed dependency, service state, or another part of the machine, a full environment restore can preserve information that those lighter-weight methods cannot.

Agentless provides a valuable counterpoint. Xia et al.\ show that a relatively simple localization, repair, and validation pipeline can compete with more elaborate autonomous agents~\citep{xia2024agentless}. We view this as an important lesson for our own work. More machinery is only worthwhile when it preserves something that matters. On the file-centric Terminal-Bench tasks we tested, independent retries remain stronger under the matched budget. On tasks with important state outside the repository, full snapshots become much more valuable.

\paragraph{Returning to promising states.}
Our search procedure is most directly inspired by Go-Explore. In this influential line of work, Ecoffet et al.\ show that exploration becomes easier when an agent can remember promising states, return to them reliably, and continue exploring from there~\citep{ecoffet2021goexplore}. Intelligent Go-Explore extends the same archive-and-return idea by using a foundation model to choose which archived state to restore from~\citep{lu2025ige}.

We bring that intuition into coding-agent environments by changing what a saved state means. In Atari, a cell represents a compact environment state. In our setting, the equivalent object is a live software sandbox containing files, installed dependencies, process state, databases, and other artifacts created during execution.

This setting introduces a new challenge. A perfectly restored state can still be a poor place to continue from. It may contain a wrong partial solution. The next agent may not understand how that state was reached. Revisiting it also consumes tokens that could have been spent on a fresh attempt. Our experiments therefore focus on whether a restored state is actually worth returning to.

\paragraph{Positioning.}
Prior work has shown strong results from preserving generated candidates, textual experience, and source-level progress. Our work adds full machine state to that design space. We do not argue that full snapshots are always better. Instead, we study when they preserve useful information that source files and textual memory miss. Our results suggest that this distinction becomes most important when progress extends beyond the repository.

\section{Conclusion}
\label{sec:conclusion}

Full sandbox snapshots let coding agents return to intermediate execution states that cannot always be represented by a patch or textual summary. The value of that capability depends on what the environment contains and whether the system can identify states that are actually worth revisiting.

For file-centric tasks, simpler representations and fresh retries are often enough. When progress lives in databases, services, installed dependencies, or other machine state outside the repository, full snapshots become more compelling.

The main open problem is selection: how can we tell when a machine state represents reusable progress rather than a dead end? Better progress signals, cheaper handoffs between agents, and tasks with richer environment state are promising directions for making environment-level search useful in practice.

\bibliography{references}
\bibliographystyle{mlsys2026}

\end{document}